\documentclass{article}

\usepackage[accepted]{icml2021}

\usepackage[utf8]{inputenc}
\usepackage[T1]{fontenc}
\usepackage{microtype}

\usepackage{bm}
\usepackage{graphicx}
\usepackage{amsmath,amsfonts}
\usepackage{booktabs}
\usepackage{multirow}
\usepackage{makecell}
\usepackage{textcomp}
\usepackage{array}
\usepackage{textcomp}
\usepackage[font=footnotesize]{subfig}
\usepackage{multirow}
\usepackage{float}
\usepackage{xspace}
\usepackage[export]{adjustbox}
\usepackage{nicefrac}
\usepackage{colortbl}
\usepackage{tablefootnote}
\usepackage[referable]{threeparttablex}
\usepackage{comment}
\usepackage{listings}

\usepackage{mathtools}

\DeclarePairedDelimiterX{\infdivx}[2]{(}{)}{%
	#1\;\delimsize\|\;#2%
}

\makeatletter
\@ifpackageloaded{icml\the\year}{
\usepackage{enumitem}
\setlist{nosep}}
{\usepackage[linesnumbered,ruled,vlined]{algorithm2e}
\SetKwRepeat{Do}{do}{while}
\DontPrintSemicolon

\SetCommentSty{commentsty}
}
\@ifclassloaded{acmart}{
\def\authornotetext#1{
	\g@addto@macro\@authornotes{%
	\stepcounter{footnote}\footnotetext{#1}}%
}}{
\usepackage[usenames,svgnames]{xcolor}
\usepackage{hyperref}
\hypersetup{
    colorlinks = true,
    breaklinks = true,
    bookmarksdepth = section,
    citecolor = DarkGreen,
    linkcolor = DarkRed,
    urlcolor = DarkBlue,
}}
\@ifclassloaded{llncs}{
\usepackage[square,comma,numbers,sort&compress]{natbib}
\bibliographystyle{splncsnat}

}{
\usepackage{amsthm}

\theoremstyle{remark}

}
\makeatother

\definecolor{sblue}{HTML}{02BCD4}
\definecolor{sred}{HTML}{F44436}
\definecolor{spink}{HTML}{E91E62}
\definecolor{sgreen}{HTML}{8BC34A}
\definecolor{spurple}{HTML}{3F51B5}
\definecolor{slightgreen}{HTML}{CCDE3A}
\definecolor{sorange}{HTML}{FE9800}
\definecolor{sgolden}{HTML}{FFC108}

\DeclareMathAlphabet{\mathsfit}{\encodingdefault}{\sfdefault}{m}{sl}
\SetMathAlphabet{\mathsfit}{bold}{\encodingdefault}{\sfdefault}{bx}{n}

\usepackage{bbm}
\usepackage{dsfont}
\usepackage{mathtools}
\usepackage{setspace}

\usepackage[hang,flushmargin]{footmisc}
\usepackage{balance}
\usepackage{flushend}

\newcommand{\ours}{BrainNNExplainer\xspace}
\newcommand{\back}{BrainNN\xspace}

\icmltitlerunning{An Interpretable GNN Framework for Brain Network based Disease Analysis}

\begin{document}
\twocolumn[
\icmltitle{\ours: An Interpretable Graph Neural Network Framework \mbox{for Brain Network based Disease Analysis}}
\icmlsetsymbol{correspondence}{\dag}
\begin{icmlauthorlist}
\icmlauthor{Hejie Cui}{emory}
\icmlauthor{Wei Dai}{emory}
\icmlauthor{Yanqiao Zhu}{cripac,ucas}
\icmlauthor{Xiaoxiao Li}{princeton}
\icmlauthor{Lifang He}{lehigh}
\icmlauthor{Carl Yang}{emory,correspondence}
\end{icmlauthorlist}

\icmlaffiliation{emory}{Department of Computer Science, Emory University}
\icmlaffiliation{cripac}{Center for Research on Intelligent Perception and Computing, Institute of Automation, Chinese Academy of Sciences}
\icmlaffiliation{ucas}{School of Artificial Intelligence, University of Chinese Academy of Sciences}
\icmlaffiliation{lehigh}{Department of Computer Science and Engineering, Lehigh University}
\icmlaffiliation{princeton}{Department of Computer Science, Princeton University}
\icmlcorrespondingauthor{Carl Yang}{j.carlyang@emory.edu}

\icmlkeywords{graph representation learning, interpretable machine learning, brain networks}

\vskip 0.3in
]
\printAffiliationsAndNotice{}

\begin{abstract}
Interpretable brain network models for disease prediction are of great value for the advancement of neuroscience. GNNs are promising to model complicated network data, but they are prone to overfitting and suffer from poor interpretability, which prevents their usage in decision-critical scenarios like healthcare. To bridge this gap, we propose \ours, an interpretable GNN framework for brain network analysis. 
It is mainly composed of two jointly learned modules: a backbone prediction model that is specifically designed for brain networks and an explanation generator that highlights disease-specific prominent brain network connections.
Extensive experimental results with visualizations on two challenging disease prediction datasets demonstrate the unique interpretability and outstanding performance of \ours. 
\end{abstract}

\begin{spacing}{1}
\section{Introduction}
Brain networks are complex graphs with anatomic regions represented as nodes and connectivities between the brain regions as links \cite{DBLP:journals/brain/MurugesanYNDWKM20}. Interpretable models on brain networks for disease prediction play an important role in understanding the biological functions of neural systems, which can be helpful in the early diagnosis of neurological disorders and facilitate neuroscience research \cite{maartensson2018stability}. Previous models on brain networks have been studied from shallow to deep ones, such as graph kernels \cite{jie2016sub}, tensor factorizations \cite{he2018boosted, liu2018multi} and convolutional neural networks \cite {kawahara2017brainnetcnn, li2020braingnn}. 

Recently, Graph Neural Networks (GNNs) attract broad interests due to their established power in different downstream tasks \cite{kipf2016semi, xu2018powerful, velivckovic2017graph, yang2020heterogeneous}. 
Compared with shallow models, GNNs are promising for brain network analysis with more powerful representation abilities to capture the sophisticated brain network structures~\cite{maron2018invariant, DBLP:conf/nips/YangZSL019, yang2020taxogan}.

However, GNNs as a family of deep learning models are prone to overfitting and lack transparency in their predictions, which prevent their usage in decision-critical applications such as disease diagnosis. Although several approaches have been proposed to explain the predictions of GNNs \cite{ying2019gnnexplainer, luo2020parameterized, DBLP:conf/nips/VuT20, yuan2020explainability}, none of them is equipped with a backbone GNN specifically designed for brain networks. 
Moreover, they do not target at disease prediction, and produce an independent explanation for each instance, whereas for brain networks, we assume that subjects having the same disease may share similar brain network patterns, which means globally shared explanations are needed across instances. 

To unleash the power of GNNs in brain network analysis and enable their interpretability, we propose \ours. It is composed of two modules: a backbone \back \cite{Zhu:2021sp} which adapts a message-passing GNN for disease prediction on brain networks (Section \ref{sec:predict}), and an explanation generator which learns a globally shared edge mask to highlight the brain network connections that are important for specific diseases (Section \ref{sec:explainer}). 
In order to improve the prediction model and its interpretability, we further propose a three-step training strategy, where the two modules are trained on the original graph or the masked graph iteratively (Section \ref{sec:explainer}).
Through experiments on two real-world brain disease datasets (i.e. HIV and Bipolar), we show that \ours can provide explanations that are verifiable based on neuroscience findings. Furthermore, both our backbone model \back and the interpretable version \ours, especially the latter, yield significant improvements over the state-of-the-art shallow and deep baselines.

\section{\ours}

\begin{figure*}[t]
	\centering
	\includegraphics[width=\linewidth]{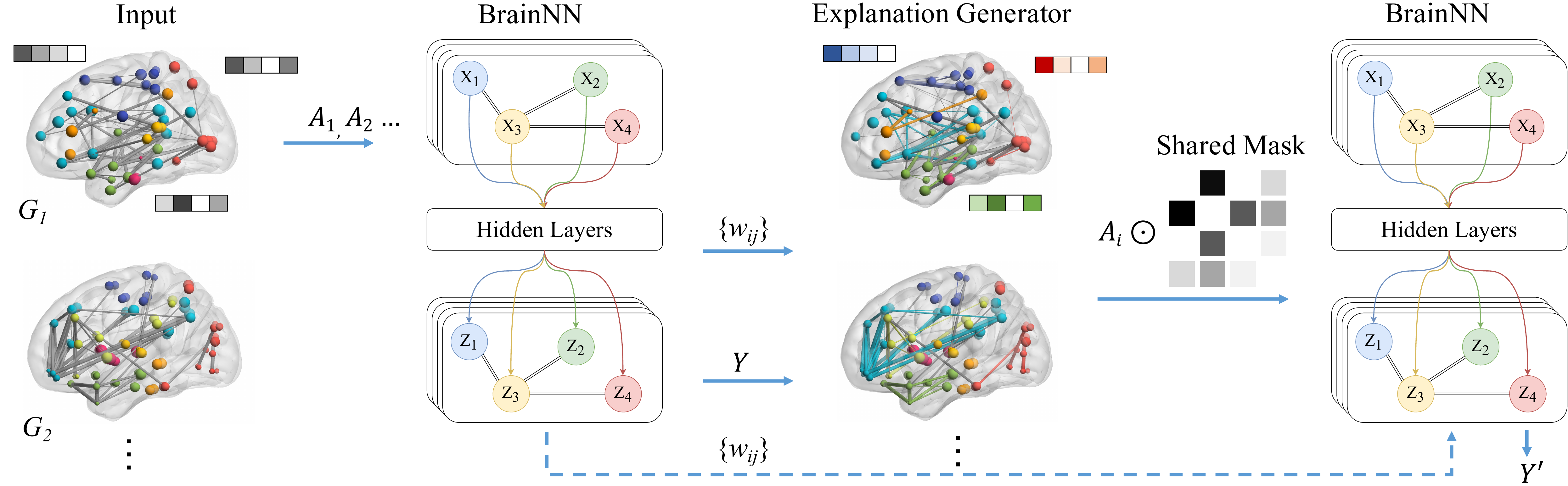}
	\caption{The proposed \ours trained in three-steps: the initial training of \back on the original data, the explanation generation based on trained \back, and the further adjustment of \back based on the explanations.}
	\label{fig:framework}
\end{figure*}

\subsection{Preliminaries}

\textbf{Problem definition.}
Given a weighted brain network \(G = (\mathcal{V}, \mathcal{E}, \bm{W})\), where \(\mathcal{V} = \{v_i\}_{i = 1}^n\) is the node set of size \(n\) defined by the regions of interest (ROIs) (same across subjects), \(\mathcal{E} = \mathcal{V} \times \mathcal{V}\) is the edge set, and \(\bm{W} \in \mathbb{R}^{n \times n}\) is the weighted adjacency matrix describing connection strengths between ROIs, the model outputs a disease prediction \(y\). We provide interpretability by learning an edge mask \(\bm{M} \in \mathbb{R}^{n\times n}\) that is shared across all subjects to highlight the disease-specific prominent ROI connections. 


\textbf{Neural system mapping.}
One unique property of brain networks is that the ROIs can be partitioned into neural systems according to their structural and functional roles under a specific atlas \cite{figley2017probabilistic, shirer2012decoding, xu2020new}, which can facilitate the verification of our generated explanations from the neuroscience perspective. 
The HIV and BP datasets we use in this paper are based on two different atlases, AAL90 and Brodmann82, respectively. We map the nodes (i.e., ROIs) on both atlases into eight commonly used neural systems, including Visual Network (VN), Auditory Network (AN), Bilateral Limbic Network (BLN), Default Mode Network (DMN), Somato-Motor Network (SMN), Subcortical Network (SN), Memory Network (MN) and Cognitive Control Network (CCN). 

\subsection{The Backbone \back}
\label{sec:predict}

\textbf{Node features construction.}
\label{sec:predict_node}
The lack of predictive original ROI features limits the power of GNNs \cite{cui2021features}. To this end, 
we construct multiple node features based on one-hot ROI identities as well as local statistical measures such as degree profiles (LDP) \cite{Cai:2018te}. In LDP, each feature \(\bm{x}_i\) of node $v_i$ is computed as
 \begin{equation}
     \small
 	\bm{x}_i = \left[ \deg(v_i); \min(\mathcal{D}_i); \max(\mathcal{D}_i); \operatorname{mean}(\mathcal{D}_i); \operatorname{std}(\mathcal{D}_i) \right],
 \end{equation}
where \(\mathcal{D}_i = \{\deg(v_j) \mid (v_i,v_j) \in \mathcal{E}\}\) describes the degree statistics of node \(v_i\)'s one-hop neighborhood, and \([\cdot ; \cdot]\) denotes concatenation. Other common artificial node features such as degree, binning degree \cite{cui2021features} and node2vec \cite{DBLP:conf/kdd/GroverL16} are also included as alternatives in our experiments. All of them are consistent across all subjects.

\textbf{Edge-weight-aware message passing.}
\label{sec:predict_mp}
Since the brain region connectivity and correlations are encoded in real-valued edge weights, which can not be handled by existing GNNs, we design an edge-weight-aware message passing mechanism. 
Specifically, we first construct a message vector \(\bm{m}_{ij} \in \mathbb{R}^{D}\) by concatenating node embeddings of a node \(i\), its neighbor \(j\), and edge weight \(w_{ij}\):
\begin{equation*}
	\bm{m}_{ij}^{(l)} = {MLP}_{\bm\Theta} \left( \left[ \bm{h}_i^{(l)};\, \bm{h}_j^{(l)};\, w_{ij} \right] \right),
\end{equation*}
where \(l\) is the index of the GNN layer. Then we aggregate messages from all neighbors followed by a non-linear transformation \cite{DBLP:conf/ijcai/XieLYW020}; the propagation rule can be written as:
\begin{equation*}
	\bm{h}_i^{(l)} = \sigma \left( \sum_{j \in \mathcal{N}_i \cup \{i\}} \bm{m}_{ij}^{(l - 1)} \right),
\end{equation*}
where \(\sigma\) is a non-linear activation function such as \(\operatorname{ReLU}\).
Finally, another MLP with residual connections is employed \cite{He:2016ib} for summarizing all node embeddings to compute graph-level embeddings \(\bm{z} \in \mathbb{R}^{D}\):
\begin{equation*}
	\bm{z}^{\prime} = \sum_{i \in \mathcal{V}} \bm{h}_{i}^{(k)}, \quad \bm{z} = {MLP} ( \bm{z}^{\prime} ) + \bm{z}^{\prime}.
\end{equation*}
This GNN can be trained w.r.t.~the supervised cross-entropy loss (denoted as $\mathcal{L}_{p}$) towards disease predictions.

\subsection{The Explanation Generator}
\label{sec:explainer}


\textbf{Shared edge mask as the explanation.}
A general approach to generate explanations for GNNs is to find a explanation graph $G^{\prime}$ that has the maximum mutual information with the label distribution $Y$, where $G^{\prime}$ can be a subgraph of $G$ \cite{ying2019gnnexplainer} or other alternations of $G$ \cite{ luo2020parameterized, schlichtkrull2020interpreting, yuan2020explainability}. Previous methods usually produce a unique explanation subgraph for each graph subject (e.g. GNNExplainer \cite{ying2019gnnexplainer}, PGExplainer \cite{luo2020parameterized}, and PGM-Explainer \cite{DBLP:conf/nips/VuT20}), or through the model-level explanation (e.g. GAT \cite{Velickovic:2018we}), that cannot drive disease-specific explanation. Considering the unique properties of brain networks (i.e. fixed number and order of nodes under a given atlas) and the characteristics of disease analysis (i.e. subjects with the same disease may share similar brain network connection patterns), 
a shared explanation graph $G^{\prime}$ is feasible in brain networks and can potentially capture more common patterns for disease-specific analysis.

To achieve this, we propose to learn a globally shared edge mask $\bm{M}\in \mathbb{R}^{n \times n}$ and apply it on the individual brain networks across all subjects in a dataset. 
Specifically, we train $\bm M$ by maximizing the mutual information between the \back predictions $\hat{y}$ on the original graph $G$ and $\hat{y^{\prime}}$ on the masked graph $G^{\prime}$, where $\bm {W}^{\prime} = \bm W \odot \sigma(\bm M)$. $\odot$ denotes element-wise multiplication, and $\sigma$ denotes the sigmoid function that maps the mask to $[0,1]^{n \times n}$. Suppose there are $C$ classes, mutual information loss can be formulated as: 
\begin{equation*}
\mathcal{L}_{m} = -\sum_{c=1}^{C} \mathds{1} [y=c] \log P_{\Phi}\left(y^{\prime}=y \mid G= \bm {W}^{\prime}\right).
\end{equation*}
The masked prediction loss $\mathcal{L}_{p^{\prime}}$ is the sum of the above mutual information loss $\mathcal{L}_m$ and the supervised disease prediction loss $\mathcal{L}_p$ from \S \ref{sec:predict}, 
$$
\mathcal{L}_{p^{\prime}} = \mathcal{L}_m + \mathcal{L}_p.
$$
We further apply a sparsity loss $\mathcal{L}_{s}$ defined as the sum of all elements of the mask parameters that imposes a regularization on the edge size of $G^{\prime}$ to obtain a compact explanation mask, and another element-wise entropy loss $\mathcal{L}_{e}$ defined as $$\mathcal{L}_{e} = -(\bm M \log (\bm M)+(1-\bm M) \log (1-\bm M))$$ from \cite{ying2019gnnexplainer} to encourage discreteness in mask weight values.

Our final training objective is
$$
\mathcal{L} = \mathcal{L}_{p^{\prime}} + \mathcal{L}_{s} +  \mathcal{L}_{e}.
$$


As a result, our explanation generator will produce a globally shared edge mask ${\bm M}$ that can highlight disease-specific prominent brain network connections, and can be further applied on all testing graphs for disease-specific neurological biomarkers and salient ROIs investigation.

\textbf{Three-step training strategy.}
Overall, \ours is trained in three steps, as shown in Figure \ref{fig:framework}. In particular, a backbone \back model is first trained on the original data, as described in \S \ref{sec:predict}. Using this trained prediction model and its prediction as the input, the explanation generator then learns a globally-shared edge mask over all training graphs with other parameters from the prediction model frozen, as described above. Finally, we apply the learned shared global mask $\bm M$ on the original training graphs $G$ to get $G^{\prime}$, then use $G^{\prime}$ to train \back backbone model again, where the parameters in backbone model will be further updated with the masked graphs. 
With this three-step strategy, we further improve the prediction model and obtain a shared explanation mask for model interpretation.


\section{Experiments}

\subsection{Datasets and Hyper-parameter Setting}
We use two real-world datasets from \cite{Ma:2017gl} to evaluate the effectiveness of our framework. For each dataset, we randomly divide 80\% for training, 10\% for validation, and the remaining 10\% for testing.

\textbf{Human Immunodeficiency Virus Infection (HIV)} is collected from functional magnetic resonance imaging (fMRI), including 35 samples from patients (positive) and 35 healthy controls (negative). Each graph contains 90 nodes (ROIs) and the edge weights corresponding to the adjacency matrix are calculated as the correlations between brain regions.

\textbf{Bipolar Disorder (BP)} is also collected from fMRI modality, consisting of 52 bipolar subjects and 45 healthy controls with matched age and gender. It stimulates 82 brain regions, according to Freesurfer-generated cortical/subcortical gray matter regions. Functional brain networks are derived using pairwise BOLD signal correlations.

\textbf{Hyper-parameter Setting.}
The proposed model is implemented using PyTorch \cite{DBLP:conf/nips/PaszkeGMLBCKLGA19} and PyTorch Geometric \cite{DBLP:journals/corr/abs-1903-02428}. We  use  Adam  optimizer \cite{DBLP:journals/corr/KingmaB14} with the initial learning rate setting to 0.001 and a weight decay of 0.00001. The backbone model, composed of three layers of multi-layer perceptron and one layer of edge-weight aware message passing  (Section \ref{sec:predict_mp}), is trained for 100 epochs with hidden dimension setting to 64. Experiments are conducted with multiple common artificial node features (Section \ref{sec:predict_node}) and different train/test/validation split. The average value of five runs under the optimal hyper-parameter setting is reported for presentation. The implementation will be available after the formal publishment of this work.

\subsection{Interpretability Analysis}
\label{sec:vis}
\textbf{Visualization.}
To qualitatively examine the effectiveness of the globally shared mask $\bm{M}$, we follow the similar strategy as the post processings in GNNExplainer \cite{ying2019gnnexplainer}, where a threshold is used to obtain a explanation subgraph $G^{\prime}_s$ by removing low-weight edges from $G^{\prime}$. 

\begin{figure}[t]
	\centering
	\subfloat[HIV]{
		\includegraphics[width=\linewidth]{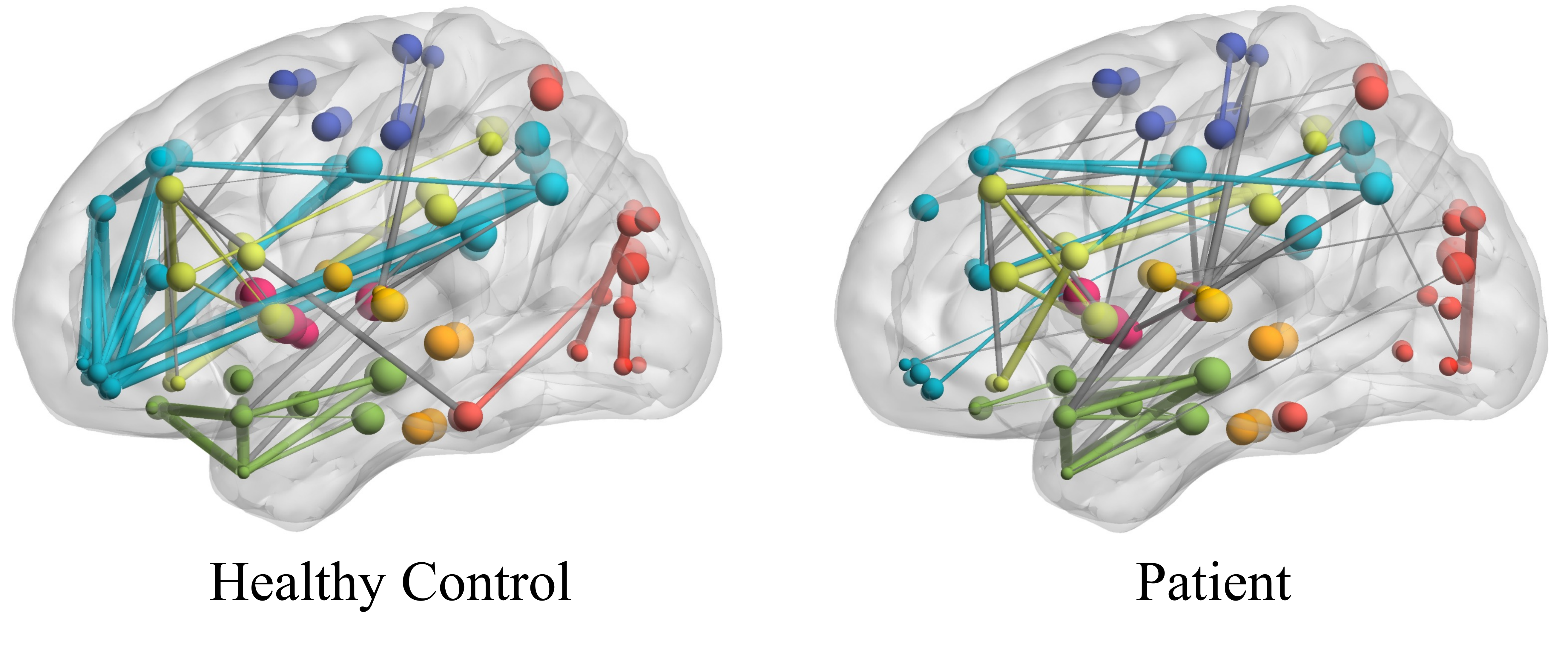}
		\label{fig:hiv_vis}
	}\\
	\vspace{-10pt}
	\subfloat[BP]{
		\includegraphics[width=\linewidth]{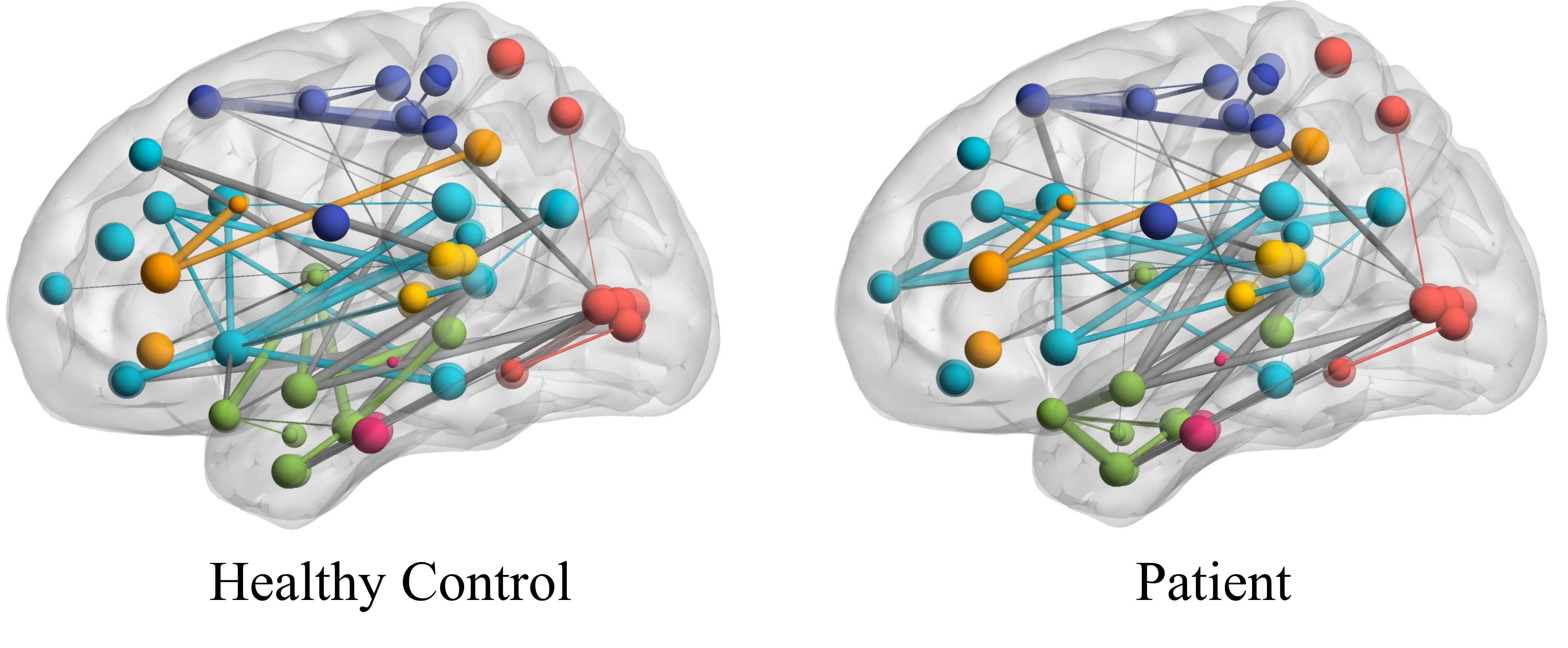}
		\label{fig:bp_vis}
	}
	\caption{Comparison of explanation graph connectomes in brain networks of a healthy control and a patient on HIV and BP datasets. The colors of neural systems are described as: {\color{sred} VN}, {\color{sgolden} AN}, {\color{sgreen} BLN}, {\color{sblue} DMN}, {\color{spurple} SMN}, {\color{spink} SN}, {\color{sorange} MN}, {\color{slightgreen} CCN}, respectively.}
	\label{fig:vis}
\end{figure}

Figure \ref{fig:vis} shows the comparison of connectomes for healthy control and patient groups on two datasets, where edges within the same systems are colored according to the color of nodes it links, while edges across systems are colored gray. The size of an edge is decided by its weight in the explanation graph. We have the following observations.

It can be seen that connectome patterns differ within certain neural systems between the healthy control and patient subject, which could provide potential value for clinical diagnosis. For example, in the HIV dataset, the explanation subgraph of patients excludes many interactions within the Default Mode Network system (DMN), which is colored {\color{sblue} blue}. The connections between superior frontal gyrus (nodes 3, 4) and orbital part (nodes 5, 6, 9, 10) are two examples\footnote{See \cite{chen2021decreased} for ROI names and respective node indices.}. Also, interactions within the Visual Network (VN, colored {\color{sred} red}) system of patients are significantly less than that of healthy controls. For example, connections between cuneus (nodes 45, 46) and lingual gyrus (nodes 47, 48), and those between occipital gyrus (nodes 51, 52, 53, 54) and fusiform gyrus (nodes 55, 56) are found to be missing. 
These patterns are consistent with the findings in \cite{herting2015default, flannery2021hiv} that alterations in within- and between-network DMN and VN connectivity may relate to known cognitive and visual processing difficulties in HIV. 

For the BP dataset, we observe that compared with tight interactions within the Bilateral Limbic Network (BLN, colored {\color{sgreen} green}) of the healthy control, the connections within BLN of the patient subject are much more sparse, which may signal pathological changes in this neural system. For instance, it is found that the patient has fewer connections between pyriform cortex (nodes 43, 44) and perirhinal cortex in the rhinal sulcus (nodes 55, 56) than healthy controls, and decreased connections between temporopolar area (nodes 61, 62) and retrosubicular area (nodes 81, 82).
These results are in line with previous studies \cite{das2020parietal, ferro2017longitudinal}. It finds that the parietal lobe, one of the major lobes in the brain roughly located at the upper back area in the skull and is in charge of processing sensory information it receives from the outside world, is mainly related to Bipolar disease attack. Since parietal lobe ROIs are contained in BLN under our parcellation, the connections missing within the BLN system in our visualization are consistent with existing clinical evidence.

\textbf{Interpretation of important brain systems.}
\begin{table}
	\centering
	\footnotesize
	\tiny
	\caption{Top ranked neural systems of the explanation subgraph on HIV and BP for both Healthy Control (Normal) and Patient under three comparative measures.}
	\begin{tabular}{ccccc}
	\toprule
	\multirow{2.5}{*}{Dataset} & \multirow{2.5}{*}{Type} & \multicolumn{3}{c}{Comparative Measures} \\
	\cmidrule(lr){3-5}
	& & Degree & Strength & Cluster Coefficient  \\
	\midrule
	\multirow{2}{*}{HIV}
	& Normal   & DMN BLN CCN & DMN BLN CCN & DMN CCN BLN  \\
	& Patient  & BLN CCN AN & BLN CCN AN & BLN  \\
	\midrule
	\multirow{2}{*}{BP}
	& Normal & BLN SMN DMN & BLN DMN SMN & SMN VN DMN  \\
	& Patient & BLN DMN SMN & BLN DMN SMN & SMN VN \\
	\bottomrule
	\end{tabular}
	\label{tab:metrics}
\end{table}
To understand which neural systems contribute most to the prediction of a specific disease, we further conduct important brain system interpretation on the explained subgraphs by observing the most manifest nodes with three commonly used measures in brain network analysis: degree, strength, and cluster coefficient \cite{DBLP:journals/neuroimage/RubinovS10}. The cluster coefficient of a node in a graph quantifies how close its neighbours are to being a clique (complete graph). Suppose the neighbourhood $\mathcal{N}_i$ for a node $v_i$ is its immediately connected neighbours $\mathcal{N}_{i}=\left\{v_{j}: e_{i j} \in \mathcal{E} \vee e_{j i} \in \mathcal{E}\right\}$ and $k_i$ is the number of neighbours of node $v_i$, the clustering coefficient for undirected graphs can be represented as 
$$
C_{i}=\frac{2\left|\left\{e_{j k}: v_{j}, v_{k} \in \mathcal{N}_{i}, e_{j k} \in \mathcal{E}\right\}\right|}{k_{i}\left(k_{i}-1\right)}.
$$

As is shown in Table \ref{tab:metrics}, important neural systems under different metrics show similar characteristics. Specifically, for HIV dataset, both healthy control and patients' explanation subgraphs reveal the importance of BLN, while DMN is missing from all three metrics in the patient group. This is consistent with our observation on HIV in Figure \ref{fig:vis}, where the densely connected structure within DMN system degenerated in patient subjects. Regarding BP dataset, BLN, SMN, and DMN are prominent in both patient and healthy controls. 

Furthermore, we compare the community structure and modularity \cite{van2010comparing} of our explanation graph $G^{\prime}$ against the original graph $G$ by conducting Newman's spectral community detection \cite{newman2013spectral}. The detected community results are compared with the ground truth neural system partition respectively with different clustering evaluation metrics. Results show that the completeness score of our explained graph achieves about 7.21\% improvement over the original graph; the Fowlkes-Mallows score improves over 5.10\%; the homogeneity score improves 5.82\%; the mutual information score improves 5.12\% and the v-measure score improves over 6.44\%. The consistent improvements of various clustering evaluation metrics validate the effectiveness of our explanation mask: after the element-wise multiplication with our trained globally-shared explanation mask, the community characteristics are further manifest than the original graphs. 






\subsection{Performance Comparison}
We compare our proposed models with baselines from both shallow and deep models for performance evaluation. 

\textbf{Metrics.}
The metrics we used in experiments to evaluate performance are Accuracy and Area Under the ROC Curve (AUC), which are both widely used measures in healthcare domain. Larger values indicate better performance.

\textbf{Baselines.}
For shallow embeddings methods, we experimented M2E \cite{Liu:2018ty}, MIC \cite{Shao:2015ek}, MPCA \cite{Lu:2008cw}, and MK-SVM \cite{Dyrba:2015ci}, where the output graph-level embeddings are further processed by logistic regression classifiers to make predictions. We also include three state-of-the-art deep models: GAT \cite{Velickovic:2019tu}, GCN \cite{Kipf:2017tc}, and DiffPool \cite{Ying:2018vl}. All the performance of baseline methods are reported under their best settings.

\textbf{Results and analysis.}
\begin{table}
\setlength{\tabcolsep}{6pt}
	\centering
	\small
	\caption{Performance of different models on HIV and BP datasets. Our methods are colored in gray background and the highest performance is highlighted in boldface.}
	\begin{tabular}{ccccc}
	\toprule
	\multirow{2.5}{*}{Method} & \multicolumn{2}{c}{HIV} & \multicolumn{2}{c}{BP} \\
	\cmidrule(lr){2-3} \cmidrule(lr){4-5}
	& Accuracy & AUC & Accuracy & AUC \\
	\midrule
	M2E   & 50.61 & 51.53 & 57.78 & 53.63 \\
	MIC   & 55.63 & 56.61 & 51.21 & 50.12 \\
	MPCA  & 67.24 & 66.92 & 56.92 & 56.86 \\
	MK-SVM & 65.71 & 68.89& 60.12 & 56.78 \\
	\midrule
	GAT   & 68.58 & 67.31 & 61.31 & 59.93 \\
	GCN   & 70.16 & 69.94 & 64.44 & 64.24 \\
	DiffPool & 71.42 & 71.08 & 62.22 & 62.54 \\
	\midrule
	\rowcolor{lightgray!20} \back & 74.29 & 71.67 & 71.11 & 64.71 \\
	\rowcolor{lightgray!20} \ours & \textbf{77.14} & \textbf{75.00} & \textbf{75.56} & \textbf{69.88}\\
	\bottomrule
	\end{tabular}
	\label{tab:performance}
\end{table}
The overall results are presented in Table \ref{tab:performance}, where our proposed backbone model \back and the prediction with three-step training from \ours (abbreviated as E-BrainNN in the table) are colored gray. Impressively, both the proposed models yield significant and consistent improvements over all SOTA shallow and deep baselines. Compared with traditional shallow models such as MK-SVM, our backbone \back outperforms them by large margins, with up to 11\% absolute improvements on BP, which demonstrates the potential of using deep GNNs on brain networks. The rationale of our edge-weight-aware message passing is supported by the superiority of \back compared with other SOTA deep models such as GAT. Based on this backbone, the performance of three-step training \ours with globally shared mask achieves a further increase of about 5\% absolute improvements. This outstanding performance of \ours certifies the unique interpretability of our explanation mask and effectiveness of the proposed framework. 
\section{Conclusion}
In this work, we propose \ours, an interpretable GNN framework for brain network based disease analysis, which consists of a brain network oriented GNN predictor and a disease analysis oriented explanation generator.
Experimental results with visualizations on two challenging disease prediction datasets validate the unique interpretability and the superior performance of our \ours. 
Under the framework of \ours, many challenges remain to be solved, such as the lack of supervision and the confinement from small scale datasets for effectively training deep GNN and explanation models. 
In the near future, we plan to conduct more ablation studies to see how much each component contributes to the system and and explore pre-training and transfer learning techniques \cite{Zhu:2021vs} based on our current pipeline. Our final aim is to build an interpretable brain analysis system eligible to digest data from different resources and modalities. 
\end{spacing}

\section*{Acknowledgements}
Lifang He was supported by NSF ONR N00014-18-1-2009 and Lehigh's accelerator grant S00010293.

\bibliographystyle{icml2021}
\bibliography{reference}
\end{document}